# Aristotle: IMO-level Automated Theorem Proving

**The Harmonic Team**

## Abstract

We introduce Aristotle, an AI system that combines formal verification with informal reasoning, achieving gold-medal-equivalent performance on the 2025 International Mathematical Olympiad problems. Aristotle integrates three main components: a Lean proof search system, an informal reasoning system that generates and formalizes lemmas, and a dedicated geometry solver. Our system demonstrates state-of-the-art performance with favorable scaling properties for automated theorem proving.

## 1 Introduction

Among the ways in which machine learning can accelerate scientific research, formal mathematical reasoning stands out as a central objective. Mathematical reasoning is foundational to most empirical sciences, is a significant bottleneck in many theoretical sciences, and is particularly well-suited for reinforcement learning. Most importantly, the task of mathematical reasoning has the potential to unlock a much broader set of reasoning abilities. This paper introduces our system, Aristotle, and describes the architecture that enabled it to achieve a gold-medal-equivalent performance on the 2025 International Mathematical Olympiad (IMO) by providing correct formal solutions to five out of the six problems [15].

We distinguish formal mathematical reasoning, by which we mean a proof written in a machine-verifiable language like Lean 4, from the informal, or natural language, reasoning that a human might write. Requiring a model to verify its reasoning in Lean 4 provides a reliable and fast reward signal across many domains of mathematics. In addition, automating the arduous manual task of formalizing mathematical arguments would make formal verification viable as a general tool for mathematicians, which could have a significant impact on mathematical research [41]. For these reasons, automated theorem proving has become a very active area of machine learning research.

Aristotle is designed to combine the strengths of formal verification with those of informal reasoning, which lets Aristotle flexibly plan and reason, in addition to leveraging the much larger corpus of mathematical proofs available in natural rather than formal language. While Aristotle may use informal reasoning and code execution to find answers and draft proofs, we only consider a problem to be solved if our system produces a complete proof using the Lean 4 proof language and its mathematical library Mathlib, without gaps or unsound axioms like `sorryAx`. We handle geometry problems separately, but still require machine-verified solutions.

Aristotle consists of three major subsystems:

1. A **Lean proof search algorithm** which ingests a Lean proof sketch and attempts to prove all unproved goals in the sketch. The search algorithm is a highly parallel Monte Carlo Graph Search (MCGS) using a large transformer as its policy and value function. The policy predicts Lean tactics conditional on the Lean proof state, proof history, and, if available, an informal proof.

2. A **lemma-based informal reasoning system** which generates informal proofs of mathematical statements, breaks these proofs down into lemmas, formalizes each lemma into Lean, and iterates this process based on formal feedback.

3. A **geometry solver** which solves plane geometry problems outside of Lean using an approach based on AlphaGeometry [45].

The version of Aristotle that was evaluated on the IMO underwent fine tuning for contest mathematics, but it was also trained broadly on many forms of formal mathematics. During training, it made novel contributions to Mathlib as well as several other open source repositories, caught subtle mistakes in a popular textbook, and showed proficiency in more advanced topics such as category theory and homological algebra; see Secion 3. This shows that Aristotle is capable of a much broader range of mathematical reasoning than that demonstrated by contest problems.

In the rest of this report we describe in more detail our main contributions:

- Developing a system, Aristotle, that reaches state-of-the-art automated theorem proving performance.
- Achieving gold-medal performance on the IMO 2025 problems.
- Demonstrating favorable scaling for automated theorem proving systems based on Monte Carlo search over possible proof steps.

There are several benchmarks for automated theorem proving for contest problems at the high school and university level [4, 46, 57, 59]. The state-of-the-art has been steadily improving on these benchmarks, with models taking various approaches, including whole-proof approaches [34, 49, 60], step-wise approaches [36, 55], and hybrid systems that combine informal reasoning, formal reasoning, and iterative refinement/correction. In addition, autoformalization/data pipelines have helped bridge informal problem statements to formal inputs [24]. Nevertheless, gold-level IMO performance has remained out of reach until this year.

ByteDance has also announced a system, Seed-Prover [5], that achieved a gold-medal IMO 2025 score with formal solutions, and OpenAI and Google Deepmind have announced gold-medal level performance with natural language solutions. We discuss these and other related systems in more detail in Section 4. Prior to this, the state-of-the-art for solving IMO-level math problems was AlphaProof and AlphaGeometry-2 [6, 12], which achieved silver-medal-level formal solutions on the IMO 2024 problems.

Our formal solutions to five of the six IMO 2025 problems are available on GitHub [14]. As of September 2025, you can register for early access to Aristotle at our website.[1]

## 2 Approach

In this section we describe in more detail the different systems that comprise Aristotle.

### 2.1 Search Algorithm

The first and most fundamental component of Aristotle is its proof search algorithm. This builds on Monte Carlo Tree Search (MCTS) with a learned value function, in the spirit of Expert Iteration [3] and AlphaZero

---

[1] https://harmonic.fun/

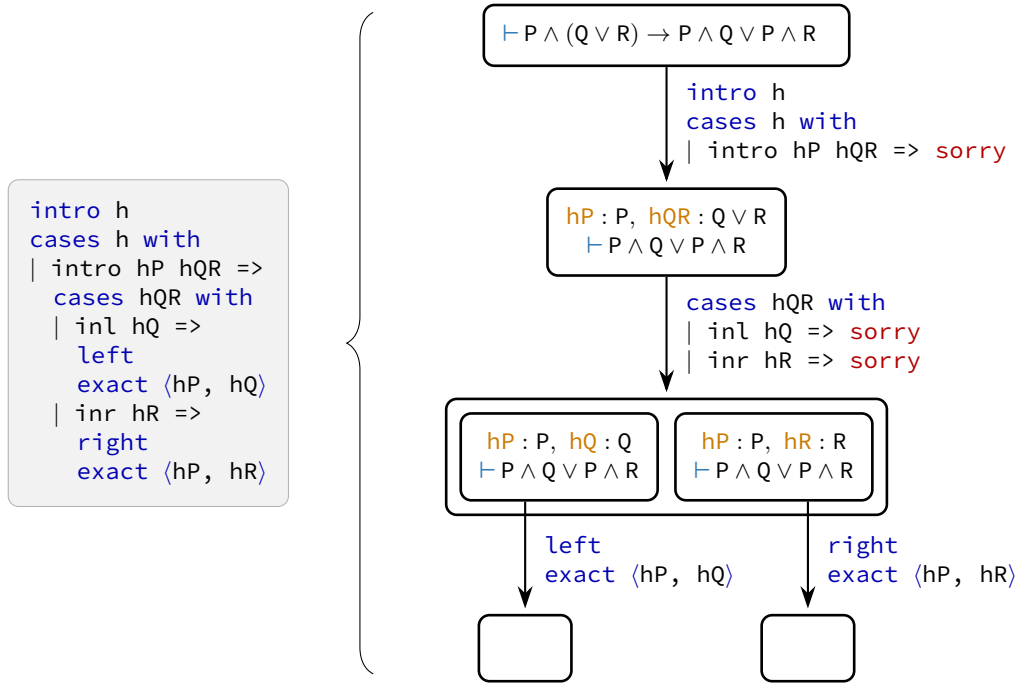

**Figure 1** A formal proof of of one direction of the distributivity of AND over OR, together with a realization as part of a search hypertree. The edges of the tree unfold the proof as a series of actions, while its nodes record how the Lean state evolves. The second `cases` tactic leads to two separate states which are resolved independently. The empty leaves indicate that no goals remain, hence that the assertion at the root is proven.

[37]. It receives a block of Lean code and attempts to replace all `sorry` statements in the code block with proofs. We use a generative policy to take progressively widened [7] samples from the large action space of Lean tactics, conditioning on the Lean proof state, proof history, and, if available, an informal proof. We use the same model and prompt (up to a task token) to compute the value function which guides the search.

The usage of a generative policy in MCTS for Lean is similar to HyperTree Proof Search [20], ABEL [11], DeepSeek-Prover-V1.5 [54], and AlphaProof [12]. The search component of Aristotle on its own is capable of proving many challenging university-level and Olympiad-level math problems, and for larger tasks it is suitable for integration into outer reasoning processes like the one described in Section 2.2, or into human-in-the-loop workflows.

### 2.1.1 States and Actions

Aristotle's search algorithm can be initiated from any collection of **Lean states**, obtained by executing an initial Lean code snippet. States are split by goals up to metavariables: if the initial code snippet or an action taken during search produces multiple goals and none of them contain metavariables, we create one state per goal. Otherwise, we retain a single state representing the collection of goals. An exception to this is when a goal is created by an explicit `sorry` placeholder; we must split these, so we do not allow such goals to have metavariables at all.

An **action** is a text string to be interpreted as a fragment of Lean code. This may consist of a single Lean tactic or a sequence of multiple tactics, and may also include informal comments. Executing an action may result in multiple states, which results in a "hypertree" as described in [20]. We can also interpret this as a tree with alternating levels by viewing the choice of state itself as a kind of action, similar to a two-player game.

When prompting our model, we include the **action history** leading to the relevant state in the prompt. While this is theoretically unnecessary (the state machine is memoryless), we found it to be helpful for reducing circular trajectories. This has the effect of allowing the model to explore a smaller set of cohesive proof strategies, rather than being forced to branch at every state in order to hit all possible proof strategies. When a reference natural language proof is available (i.e. when using our search algorithm for proof autoformalization), we also include this reference proof in the prompt.

### 2.1.2 Equivalences and Graph Search

It is natural and computationally efficient to identify states which are only superficially distinct, likewise for actions. We consider Lean states to be equivalent if they are equal in goal expressions, local context expressions, and local variable names. We consider actions to be equivalent if the state transitions they produce are equivalent. While it is not strictly true that changing the initial state of an action up to equivalence only changes its resulting states up to equivalence (a common counterexample is the aesop tactic, which uses global state internally), we are able to work around this in practice.

An effect of imposing these equivalences is to turn the search hypertree into a search hypergraph. Thus our search algorithm is more precisely a form of Monte Carlo Graph Search, investigated in the literature in [8, 50]. If one views the selection and backpropagation phases of MCTS as depth-first search and dynamic programming, respectively, then MCTS itself works on graphs. While its worst-case per-step complexity degrades from $O(D)$ on trees to $O(V)$ on graphs, where $D$ is the diameter of the graph and $V$ its number of vertices, we do not observe any CPU bottleneck in our implementation for graphs that arise in practice.

### 2.1.3 Search Strategy

Our search algorithm uses a variant of the PUCT (Predictor Upper Confidence bound applied to Trees) formula [35, 37], where the exploration bonus is weighted by a prior policy. We approximate our prior policy by the empirical distribution of actions sampled from our generative model. We avoid using sequence logprobs, which are biased in the sense of penalizing actions that have multiple equivalent textual representations.

Recall that actions may lead to multiple states, for example from tactics such as cases or code with sorry placeholders. Actions and states differ in the following respect: a state is proven if any single action succeeds (an OR condition), while an action is successful only if all of its resulting states are proven (an AND condition). This AND/OR structure makes finding a proof equivalent to a minimax problem. Accordingly, we explore actions with the highest potential (highest upper-confidence bound) and then prioritize the most challenging of their resulting states (lowest lower-confidence bound), effectively targeting the proof's bottleneck first. Furthermore, in order to prune states and disprove statements, we augment each single-goal state with a state transition corresponding to the logical negation of that goal, to which the search algorithm can allocate search budget.

### 2.1.4 Interleaving Informal Reasoning

Our model produces two kinds of informal output in addition to its formal output. First, the actions it generates consist of informal comments in addition to Lean tactics. Second, it uses a hidden chain of thought with a dynamically set thinking budget before predicting an action. The informal comments complement the chain-of-thought reasoning by allowing the model visibility into its earlier thinking later in the proof, as the hidden chain of thought is not subsequently accessible.

All three kinds of output, hidden chain of thought, informal comments, and formal Lean code, are co-evolved during training. While early in training we enforce comment production by hand, later the RL signal alone is enough to result in ubiquitous comments in the policy model's output. This confirms that this source of informal reasoning genuinely enhances the formal search. The use of a dynamic thinking budget provides a signal for when thinking traces are genuinely important for proving a given goal, which at training time we use to ensure we only train on useful traces.

### 2.1.5 Postprocessing

It is often the case that we can discover more efficient proofs by applying linter suggestions or skipping sequences of actions. A proof is an acyclic subgraph of the search hypergraph with a single source state, only successful actions, exactly one action per state, and all resulting states per action. A state may be the source of multiple proofs, with differing costs. After a search concludes, we apply linter suggestions, clean up the tactic code, and do further offline computation to condense and simplify proofs.

### 2.1.6 Reinforcement Learning

As mentioned above, we use a single model for both the value function and for action generation. We begin with a pretrained large language model and train it via reinforcement learning in the style of expert iteration [3]. Our training makes use of a large dataset of mathematical statements, including both open-source collections and in-house data. As this dataset is too large to formalize manually, we developed a statement autoformalization system, which consists of initial autoformalization, judging using signals from the Lean REPL, and correction.

We train the generative policy on proofs found by search, filtered by measures of nontriviality. We train the value function on proven states within these proofs and on nearby states that are disproven or unproven after significant effort. For statements where the dataset includes an informal proof, we also use a judge to predict the faithfulness of the formal proof, filtering out ones which are unaligned to the informal proof. Similar to Hindsight Experience Replay [2], we also render proofs of non-root states (i.e. not the main statement) as if they were root states (i.e. the main statement).

### 2.1.7 Test-Time Training

When deploying Aristotle at large scales we also use a form of test-time training (TTT) [1] to allow it to learn from its own experience at inference time. This consists of iteratively applying the following alternating steps:

1. Make some number of attempts to solve the main problem, as well as the many lemmas generated by the outer reasoning loop described in Section 2.2.
2. If the main problem remains unsolved, retrain the model on search traces extracted from these attempts.

We observed both before and during the IMO that TTT improved both the efficiency and specialization of Aristotle's attempts on a given problem. Prior to the IMO, we found that using TTT allowed Aristotle to solve difficult problems that were unsolved by our base model at the same search budget. TTT also interacts especially constructively with Aristotle's lemma generation pipeline, allowing cross-pollination between lemmas from different proof sketches, and accelerating the rate at which Aristotle processed lemmas during our IMO evaluation.

In another direction, TTT helps in working with new Lean abstractions (or less commonly seen abstractions from Mathlib) which have only basic APIs. Proving statements which involve these requires synthesizing higher level constructions in an ad hoc way from complex chains of basic commands. TTT is invaluable in allowing the model to learn from its initial attempts at chaining together basic commands in this way, effectively giving it access to a more robust API in later attempts.

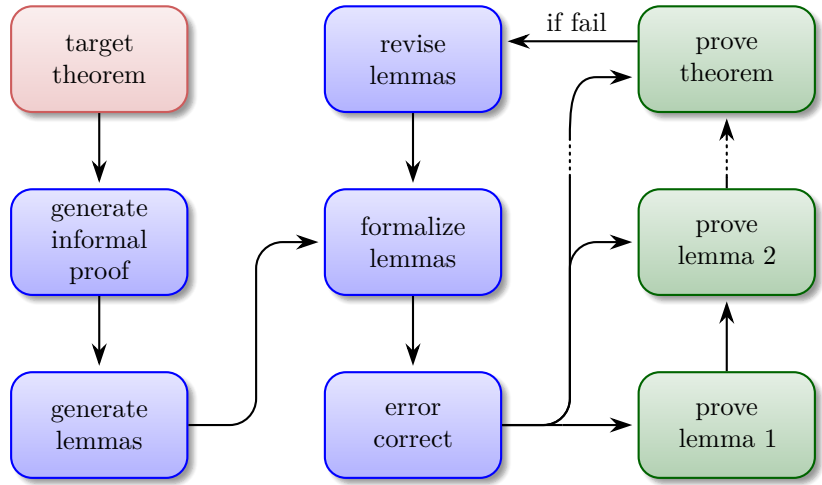

**Figure 2** The target theorem (red) is processed through a pipeline of natural language queries (blue), resulting in a series of lemmas which are processed by the search algorithm (green). If the theorem remains unproven at the end, the unproven lemmas are revised and the procedure is repeated.

## 2.2 Lemma-based Reasoning

While the search algorithm described in the previous section is effective by itself, it becomes more powerful when deployed as part of a larger lemma-based reasoning system.

### 2.2.1 Lemma Generation Pipeline

As discussed above, Aristotle's search algorithm can be initialized with any existing Lean code block. Crucially, this code block can contain already proven background results or lemmas tailored to the target theorem, and the policy model will be able to leverage these.

This feature makes it possible to boost the model's performance using higher-level informal reasoning. Namely, we can generate a list of candidate lemmas that may be useful for the target statement, and have the search algorithm attempt these first before attempting the target statement. An early implementation of this strategy is the Draft, Sketch, and Prove algorithm of [19], and many variations have since appeared in the literature.

Our implementation of this strategy is summarized in Figure 1. Beginning with the target theorem, we process it through a pipeline of natural language queries:

1. First, we ask for an informal proof of the theorem.

2. Next, we ask for the proof from (1) to be restructured as a sequence of lemmas which build on each other, and which individually have very short proofs.

3. We then ask for formalizations of the statements of the lemmas produced in (2).

4. Finally, after sending the formalizations from (3) to the Lean REPL, we communicate any error messages back and ask for corrections.

More precisely, each of these steps consists of multiple subqueries which give the model a chance to reflect on its initial output and edit it with suggested considerations in mind.

Of course, this is a noisy process: typically there will be both gaps in the proposed informal proof and errors in the formalizations of the lemma statements, even after the error correction in (4). However, in practice it produces enough useful and correctly formalized lemmas to meaningfully improve performance over the baseline of the search algorithm alone.

### 2.2.2 Iteration with Formal Feedback

This lemma architecture can also be applied iteratively if Aristotle still fails to prove the target theorem. Specifically, we consider the following iterative framework. At the end of a failed attempt, some lemmas will generally have been proven while others will remain unproven (along with the target theorem). We annotate the list of lemmas used for the attempt by marking each as proved or unproved, then use this as input for a modified version of the original query sequence:

1. First, we ask for a revision of the previous lemma list, keeping the proved lemmas but complementing them with a new collection of additional lemmas.

2. We then ask for formalizations of the statements of the new lemmas produced in (1).

3. After sending the new formalizations from (2) to the Lean REPL, we again communicate any error messages back and ask for corrections.

Broadly speaking, the original attempt will have failed for a combination of the following two reasons. First, the original proof, or at least the portion that was not proved, may follow an unworkable strategy which needs to be fundamentally altered. And second, the strategy may have been sound but not broken down into sufficiently granular lemmas for the search algorithm to prove them directly. The prompt for step (1) instructs that both possibilities should be considered in the revision.

## 2.3 Geometry

Our geometry solver is based on Yuclid, a very fast C++ DD/AR (deductive database and algebraic reasoning) engine. We have published Yuclid under Apache 2.0 on Github, and we look forward to seeing how the open source community builds with it. All parts of Yuclid are written with speed optimization in mind, making it up to 500x times faster than AlphaGeometry-1.

### 2.3.1 Performance Optimizations

Here are some optimizations we use in Yuclid.

**Numerical rule matcher** Before running the main DD/AR loop, Yuclid finds all the rules that match numerically, i.e. whose hypotheses and conclusions are numerically true. To avoid redundant effort later, the matcher finds certain configurations in the diagram (e.g., pairs of similar triangles, midpoints, triangle bisectors etc), then adds all rules about these configurations to the list.

**Deduplicating statements** In the main DD/AR loop, Yuclid goes over all numerically matched rules and tries to deduce their hypotheses and conclusions from AR. In order to do that, it stores the current progress towards proving statements in a dictionary, thus deduplicating effort between different occurrences of the same statement.

**AR optimizations** The algebraic reasoning is implemented using Gaussian elimination, with a few optimizations allowing us to avoid repeated work. For one, we store the current echelon form of the linear system, updating it every time we establish a new statement. For each statement we attempt to prove by AR, we also store its current form reduced modulo already established equations. This way, the next time we try to prove the same statement, we can resume work instead of restarting it.

**Memory management** The features of the C++ language, as well as STL and Boost containers, allow us to optimize memory management. For instance, we store the reduced forms of all pending AR equations in a dictionary, and the corresponding statements store raw pointers to the objects in this dictionary instead of, e.g., keys.

As a result of these optimizations, the main DD/AR loop takes about the same time as the numerical rule matcher, and the whole program runs faster than building a numerical diagram in Python. With prebuilt numerical diagrams, Yuclid solves 17 out of 30 problems in the AG-30 set, and saturates the rest, in about 0.4s while running on a single 3.1GHz core.

### 2.3.2 Other Features

Besides performance optimizations, Yuclid introduces several extensions compared to AlphaGeometry-1. The set of extensions is mostly disjoint with those introduced in AlphaGeometry-2. In future versions we will also incorporate some of these AG-2 extensions into our system.

**New AR table** In addition to the AR tables for the lengths and for the logarithms of lengths, Yuclid introduces a new AR table for squared lengths. This table allows us to efficiently state the Pythagorean Theorem, as well as its more general form

$$AB \perp CD \iff AC^2 + BD^2 = AD^2 + BC^2.$$

**Law of sines** Optionally, Yuclid can add sines of the angles to the ratios AR table, together with the law of sines. This extension allows it to prove one new problem from AG-30 by DD/AR, but also slows down Yuclid by approximately 10x so is turned off in our recommended configuration.

**Equations as statements** In Yuclid, any AR equation in either of the three AR tables can be wrapped into a Statement. This way we can add, e.g., the parallelogram law without adding rule-specific predicates.

**New rules** After running data generation for the first time, we searched the output for problems with up to 4 points that needed auxiliary constructions, turning several of these into new rules for subsequent use.

## 3 Results

In this section we describe in more detail our evaluation of Aristotle on the 2025 IMO, and present some other results obtained during Aristotle's development which we found interesting.

### 3.1 IMO 2025

Each annual IMO consists of six problems drawn from algebra, number theory, geometry, and combinatorics. In the 2025 IMO, human contestants were awarded a gold medal if they achieved a perfect score on five of these. Aristotle produced a formally verified solution for all but the final problem, widely regarded as the most difficult and also unsolved by the other AI systems which attained gold-medal-equivalent performance.

For each of the problems, we translated the given informal statement by hand into a formal statement. We emphasize that this was the only human formalization step in our evaluation process, with the statements of the many lemmas appearing in Aristotle's proofs being the result of autoformalization. For statements requiring contestants to first identify an unknown (e.g. the constant c = 4 appearing in Problem 3), solution candidates were first generated by a language model with access to basic mathematical programming tools.

Obtaining the maximal performance of our system required scaling it in three orthogonal ways:

1. Running our search algorithm with a model having over 200B parameters.

2. Running many parallel instances of the lemma-based reasoning pipeline to sample widely from the space of informal proofs and formal sketches.

3. Iterating the formal feedback loop of this pipeline multiple times on individual instances to maximize the potential for error correction.

As discussed in Section 2.1.7, we also used test-time training to maximize the performance returns of this scaling.

Below we provide some further analysis of Aristotle's solutions to the four non-geometry problems it solved. We aim both to highlight the most interesting aspects of the resulting Lean code, and to use these examples to illustrate some compelling features of Aristotle's capabilities.

### 3.1.1 Problem 1: Sunny Lines

The proof produced for "Sunny Lines" was unlike most human proofs we saw. This problem concerned lines passing through a number of lattice points within a triangle, and a typical step of the proof uses the fact that a line can pass through at most two edges of a convex body (here, a triangle). This fact is geometrically obvious but cumbersome to formulate and justify formally in Lean, and instead Aristotle produced a different, more algebraic proof that studied certain derangements (permutations without fixed points) which implicitly appear in the problem. It found an explicit counterexample by examining points up to $n = 5$, ruling out the construction for all greater $n$.

### 3.1.2 Problem 3: Bonza Functions

The "Bonza functions" problem required characterizing functions $f : \mathbb{N} \to \mathbb{N}$ obeying the congruence $b^a \equiv f(b)^{f(a)} \pmod{f(a)}$. In the early stages of problem solving, Aristotle started to characterize these functions in terms of a certain set S, which it ultimately formalized as

$$\texttt{def S (f : } \mathbb{N}_+ \to \mathbb{N}_+\texttt{) : Set } \mathbb{N}_+\texttt{:= \{p | Nat.Prime (p : } \mathbb{N}\texttt{)} \land \texttt{ f (p) > 1\}}.$$

This set was used in the final proofs of several lemmas, but was not directly defined or suggested by the problem statement. Thus it illustrates Aristotle's ability to define and use novel auxiliary definitions, much as a human mathematician would.

### 3.1.3 Problem 4: Sums of Three Divisors

Problem 4 concerns when it is possible to have an infinite sequence of numbers obeying a certain recurrence relation involving their proper divisors. The argument that Aristotle developed uses much the same logic as a typical human solution – there is essentially just one argument for it – but the process leading to the proof contained several illustrative examples of the power of formalization for identifying and correcting errors in informal reasoning.

Consider the following excerpt from one of Aristotle's informal proofs which led to the ultimate formal proof:

> ... in this case, $a_{n+1} < a_n$. If the sequence enters such a state, it becomes strictly decreasing. For an infinite sequence of positive integers, this is only possible if it eventually becomes constant, i.e., it reaches a fixed point.

Here the informal reasoning becomes confused over the distinction between strictly decreasing sequences, weakly decreasing sequences, and sequences which decrease at some point—all three of which it refers to as "decreasing sequences" in the surrounding text. This kind of subtle language error is emblematic of the challenges of verifying AI-generated natural language proofs.

Nonetheless, during the subsequent formalization process Aristotle repaired this error (and some related ones), using the key ideas from the flawed informal proof but filling its logical gaps as necessary. Moreover, since the accompanying informal comments in the final proof follow the validated structure of the Lean code, they provide a corrected informal account of the ultimate logic.

### 3.1.4 Problem 5: Inekoalaty Game

The solution to the "Inekoalaty Game" required identifying optimal play strategies for two players in a certain game, and proving under what conditions they force wins and draws. Like Problem 3, Aristotle identified the appropriate play strategies and formalized their definitions in Lean, but more interestingly introduced auxiliary definitions relevant to the proof but not suggested by the statement itself. Specifically, it defined the function $f(k) := k\sqrt{2}/(2k-1)$ which turns out to become a useful invariant during the proof process.

In this problem Aristotle also displayed its fluency with topics more advanced than necessary for IMO problems, which in principle can be solved using only concepts from high school mathematics. Its initial proof of the lemma `lemBazzaObjectiveUpperBound` used calculus to obtain a certain useful upper bound from the maxima of another auxiliary quantity, $\Phi(k,q) = \sqrt{(k-1)q} + \sqrt{2k-q}$, computed from its derivatives in $q$. Later, however, it replaced this calculus-based proof with a more direct proof using the `nlinarith` tactic.

Another instance was its surprising use of the `filter_upwards` tactic. This tactic is used in the manipulation of filters, typically absent from an undergraduate math curriculum, but used in the underlying definitions of limits in Mathlib. Aristotle used this tactic to prove that the quantity $f(k)$ above approaches $1/\sqrt{2}$. This can be done in other ways, but the use of such filter manipulations illustrates Aristotle's ability to deploy a broad range of techniques, including relatively obscure ones.

## 3.2 Other Results

Beyond its IMO performance, Aristotle has already demonstrated usefulness in some less artificial mathematical contexts. During training, Aristotle proved several important theorems missing from Mathlib, including Niven's theorem [28], the Gauss-Lucas theorem, the fact that eigenvalues are the roots of the characteristic polynomial [26], and other technical lemmas [27].

Aristotle also solved problems involving mathematics beyond the competition or undergraduate level. As an illustration, we include in Appendix C two instances of it proving basic statements involving homological algebra and Eisenstein series, respectively. These anticipate the broader mathematical ability we are building into future versions of Aristotle as we develop it into a working mathematical research assistant.

Aristotle has also been able to contribute to other repositories besides Mathlib. Aristotle contributed two lemmas [25] to the Polynomial Freiman-Ruzsa formalization project [42]. It has also contributed several lemmas to the Generalized Quantum Stein's Lemma project [29].

Finally, Aristotle was able to validate significant parts of Terence Tao's Lean-based real analysis textbook [38, 39]. Upon processing a few chapters from the book, Aristotle discovered four exercises which were false as written, providing explicit counterexamples. Aristotle was able to prove all other exercises it was presented with, and for two of them noticed that one of the hypotheses given to the students was unnecessary to complete the proof. See Appendix D for these examples. This textbook uses its own pedagogical definition of integers and sets, distinct from Mathlib and Lean core, which moreover shows Aristotle's ability to handle novel types and work with them over long contexts.

## 4 Related Work

ByteDance recently announced Seed-Prover [5], a system that also achieves gold medal performance with formal solutions on IMO 2025. Some common elements it shares with our system are the use of a specialized subsystem for geometry problems, the combination of informal reasoning with formal feedback, the decomposition of problems into lemmas and conjectures, and the use of solved lemmas as context for subsequent reasoning. This convergent evolution suggests that the use of natural language reasoning to decompose problems into simpler subproblems, combined with RL based on formal feedback from a Lean compiler, is a crucial tool for achieving high levels of formal mathematical reasoning.

Among the various differences between Seed-Prover and Aristotle, the most fundamental is that Seed-Prover generates a whole proof for a given formal statement and then iteratively refines it until it is verified by a Lean compiler, whereas Aristotle employs an informal reasoning model to guide a tree search as it develops a proof one step at a time. It is noteworthy that both approaches reached gold-medal-level IMO performance at the same time. Indeed, over the last two years, the leader on other automated theorem proving benchmarks has gone back and forth between step-wise and whole proof systems.

Google Deepmind and OpenAI also achieved gold medal performance on IMO 2025, but their systems provided natural language solutions instead of formal solutions and were evaluated by human judges. OpenAI used a general purpose reasoning model, and Deepmind used an advanced version of Gemini Deep Think that was finetuned for mathematical reasoning. These are impressive demonstrations of the reasoning capabilities of frontier models. However, verifying a complicated natural language proof can be almost as challenging for a human as creating a proof, so we continue to believe that the development of formally verified automated mathematical reasoning is central to potential applications in mathematical and scientific research.

We have mentioned some of the most important antecedents we build on throughout the text, but there is by now a large literature on deep learning for automated theorem proving. We refer to [9, 10, 13, 17, 18, 21–23, 30–33, 43, 47, 48, 51–53, 56, 58, 61] for a broader overview of the area.

## 5 Conclusion

In this paper, we presented Aristotle, a system that achieved a gold-medal-level performance on the 2025 IMO problems by successfully integrating informal, human-like reasoning with the rigor of formal verification in Lean 4. Our approach utilizes a sophisticated proof search algorithm to complete formal proofs, integrating a system that generates and formalizes high-level lemmas, demonstrating a successful synergy between these two reasoning levels. Beyond its success in competitive mathematics, Aristotle's ability to contribute to Mathlib and reason about advanced mathematical topics shows that this hybrid approach is a promising step toward creating AI systems that can act as valuable collaborators in mathematical research. These results suggest that this combination of large-scale models, guided search, and formal verification provides a robust path toward increasingly sophisticated automated reasoning and discovery.

## A  Contributors and Acknowledgements

The Harmonic team would like to give a special thanks to: Dagur Ásgeirsson, Rida Hamadani, Chris Hughes, John Jennings, Lorenzo Luccioli, Pietro Monticone, Daniel Morrison, Albert Smith, Damiano Testa, Edward Watine, Kexing Ying, Jujian Zhang.

Finally, we would like to extend our gratitude to the Lean community, whose contributions have been foundational to Aristotle.

## B  Lean Interaction

We interact with Lean through a REPL. The REPL is a program written in Lean itself (based on [44]) that manages Lean goal states and can apply Lean tactics to them as if they were part of a larger Lean file. The REPL supports initializing from a Lean file with `sorrys` in producing a goal state for each sorry, and running Lean tactics on individual proof states.

When a proof is found via the search process, it is verified by rendering the proof out as a self contained Lean file and running it through Lean as a command in order to check for kernel errors, check which axioms are used, and check which lemmas from the file were necessary in order to complete the proof. The REPL also allows us to apply more stringent checks to ensure that the proof terms produced by tactics are well typed. This allows us to eagerly catch errors that would otherwise only be found by the kernel at the end of the proof.

The REPL is hosted on many CPU-only machines, which operate in a stateless manner. Any REPL process can be asked to apply a tactic at some state at any time, as all the information required to reconstruct the state is contained within the request. The machines are hosted in GKE which can be scaled up and down depending on demand and can be routed to on a per-request basis. For more details on this system and a discussion of its development, see [16].

## C  Example Proofs

As mentioned in Section 3, we showcase here two selected proofs demonstrating Aristotle's ability to work with more abstract or advanced concepts than those relevant to competition mathematics. The first is the following exercise in homological algebra.

```
/-
Let $$0 \to A_n \to B_n \to C_n \to 0$$ and $$0 \to A_{n+1} \to B_{n+1} \to C_{n+1} \to
    0$$ be two short exact sequences of modules (or abelian groups) forming a
    commutative diagram: $$\begin{array}{ccccccc} 0 & \to & A_n & \to & B_n & \to & C_n
    & \to & 0 \\ & & \downarrow & & \downarrow & & \downarrow & \\ 0 & \to & A_{n+1} &
    \to & B_{n+1} & \to & C_{n+1} & \to & 0. \end{array}$$ Assume that the vertical
    arrows $$A_n \to A_{n+1}$$, $$B_n \to B_{n+1}$$, and $$C_n \to C_{n+1}$$ are
    morphisms. If the morphisms $$A_n \to A_{n+1}$$ and $$C_n \to C_{n+1}$$ are
    isomorphisms, prove that the morphism $$B_n \to B_{n+1}$$ is also an isomorphism.
-/
theorem formal_2068504 (A B C : ℕ → Type _) [∀ n, (AddCommGroup (A n))] [∀ n, (
    AddCommGroup (B n))] [∀ n, (AddCommGroup (C n))]
    (f : ∀ n, ((A n) →+ (B n)))
    (g : ∀ n, ((B n) →+ (C n)))
    (h : ∀ n, ((A n) →+ (A (n + 1))))
    (k : ∀ n, ((B n) →+ (B (n + 1))))
    (l : ∀ n, ((C n) →+ (C (n + 1))))
    (hex_n : ∀ n, ((f n).range = (g n).ker))
    (hinj_f_n : ∀ n, (Function.Injective (f n)))
    (hsurj_g_n : ∀ n, (Function.Surjective (g n)))
```

```
      (hcomm_fg_k : ∀ n, (((k n).comp (f n)) = ((f (n + 1)).comp (h n))))
      (hcomm_gk_l : ∀ n, (((l n).comp (g n)) = ((g (n + 1)).comp (k n))))
      (hiso_h_n : ∀ n, (Function.Bijective (h n)))
      (hiso_l_n : ∀ n, (Function.Bijective (l n))) : ∀ n, (Function.Bijective (k n)) :=
    by
      -- First, we show that $k$ is injective.
      have hinj_k_n : ∀ n, (Function.Injective (k n)) := by
        -- Assume that $k_n(x) = k_n(y)$ for some $x, y \in B_n$.
        intro n
        by_contra h_k_not_inj;
        -- Since $k_n$ is not injective, there exists $x \in B_n$ such that $k_n(x) = 0
$ and $x \neq 0$.
        obtain ⟨x, hx_ker, hx_ne_zero⟩ : ∃ x : B n, k n x = 0 ∧ x ≠ 0 := by
          contrapose! h_k_not_inj;
          intro x y hxy;
          exact sub_eq_zero.mp ( h_k_not_inj ( x - y ) ( by simp ( config := { decide
:= Bool.true } ) [ hxy ] ) );
        -- Since $l_n$ is an isomorphism, we have $g_n(x) = 0$.
        have hg_n_x_zero : g n x = 0 := by
          replace hcomm_gk_l := congr_arg ( fun f => f x ) ( hcomm_gk_l n );
          aesop;
          exact ( hiso_l_n n ).1 ( by simpa using hcomm_gk_l );
        -- Since $g_n(x) = 0$, there exists $a \in A_n$ such that $f_n(a) = x$.
        obtain ⟨a, ha⟩ : ∃ a : A n, f n a = x := by
          -- Since $g_n(x) = 0$, we have $x \in \ker(g_n)$.
          have hx_ker_g_n : x ∈ AddMonoidHom.ker (g n) := by
            exact hg_n_x_zero;
          -- Since $x \in \ker(g_n)$ and $\range(f_n) = \ker(g_n)$, we have $x \in \
range(f_n)$.
          have hx_range_f_n : x ∈ AddMonoidHom.range (f n) := by
            exact hex_n n ▸ hx_ker_g_n;
          exact hx_range_f_n;
        -- Since $f_{n+1}$ is injective and $f_{n+1}(h_n(a)) = 0$, it follows that $h_n
(a) = 0$.
        have hh_n_a_zero : h n a = 0 := by
          replace hcomm_fg_k := congr_arg ( fun H => H a ) ( hcomm_fg_k n );
          aesop;
        have := hiso_h_n n;
        have := this.injective;
        have := @this a 0; simp_all ( config := { decide := Bool.true } ) ;
      refine fun n => ⟨ hinj_k_n n, ?_ ⟩;
      -- Since $g_{n+1}$ is surjective, for any $x \in B_{n+1}$, there exists $y \in
B_n$ such that $g_{n+1}(k_n(y)) = g_{n+1}(x)$.
      have h_surjective_step : ∀ x : B (n + 1), ∃ y : B n, g (n + 1) (k n y) = g (n +
1) x := by
        -- Since $l_n$ is surjective, for any $x \in B_{n+1}$, there exists $y \in B_n$
 such that $l_n(g_n(y)) = g_{n+1}(x)$.
        have h_surjective_step : ∀ x : B (n + 1), ∃ y : B n, l n (g n y) = g (n + 1) x
:= by
          exact fun x => by rcases hiso_l_n n |>.2 ( g ( n + 1 ) x ) with ⟨ y, hy ⟩ ;
exact ⟨ Classical.choose ( hsurj_g_n n y ), by rw [ ← hy, Classical.choose_spec (
hsurj_g_n n y ) ] ⟩ ;
        aesop;
```

```
        obtain ⟨ y, hy ⟩ := h_surjective_step x;
        use y;
        -- By the commutativity of the diagram, we have $l_n(g_n(y)) = g_{n+1}(k_n(y))$
    .
        have h_comm : l n (g n y) = g (n + 1) (k n y) := by
          exact FunLike.congr_fun ( hcomm_gk_l n ) y;
        exact h_comm.symm.trans hy;
      -- Let x be an arbitrary element of B_{n+1}. By h_surjective_step, there exists y
    in B_n such that g_{n+1}(k_n(y)) = g_{n+1}(x).
        intro x
        obtain ⟨y, hy⟩ := h_surjective_step x;
        -- Since $k n y - x \in \ker(g (n + 1))$, there exists $a \in A (n + 1)$ such
    that $f (n + 1) a = k n y - x$.
        obtain ⟨a, ha⟩ : ∃ a : A (n + 1), f (n + 1) a = k n y - x := by
          -- Since $k n y - x \in \ker(g (n + 1))$, there exists $a \in A (n + 1)$ such
    that $f (n + 1) a = k n y - x$ by definition of the kernel.
          have h_ker : k n y - x ∈ AddMonoidHom.ker (g (n + 1)) := by
            simp ( config := { decide := Bool.true } ) [ AddMonoidHom.mem_ker, hy ];
          rw [ ← hex_n ] at h_ker;
          exact h_ker;
        -- Since $h_n$ is bijective, there exists $b \in A_n$ such that $h_n(b) = a$.
        obtain ⟨b, hb⟩ : ∃ b : A n, h n b = a := by
          exact hiso_h_n n |>.2 a;
        use y - f n b;
        replace hcomm_fg_k := congr_arg ( fun z => z b ) ( hcomm_fg_k n ) ; aesop
```

The second establishes the following basic fact about Eisenstein series.

```
/-
Let $g_2(\\tau)$ and $g_3(\\tau)$ be the Eisenstein series associated with the lattice
    $\\tau
eq 0$, and let $q = e^{2\\text{pi}i\\tau}$ denote the modular parameter. Define the
    discriminant $\\text{Delta}$ as $\\text{Delta} = g_2^3 - 27g_3^2$. Prove that under
    modular transformations $\\tau \\to -1/\\tau$, the discriminant satisfies the
    transformation property: $\\text{Delta}(-1/\\tau) = (\\tau/i)^{12} \\text{Delta}(\\
    tau)$.
-/
theorem formal_2210385 (g₂ g₃ Δ : ℂ → ℂ)
    (h₀ : ∀ τ, ((g₂ τ) = (2 / 3) * ∑' (x : ℤ × ℤ), if ¬(x.1 = 0 ∧ x.2 = 0) then ((τ *
    x.2 + x.1) ^ 4)⁻¹ else 0)))
    (h₁ : ∀ τ, ((g₃ τ) = (1 / 27) * ∑' (x : ℤ × ℤ), if ¬(x.1 = 0 ∧ x.2 = 0) then ((τ *
    x.2 + x.1) ^ 6)⁻¹ else 0)))
    (h₂ : ∀ τ, ((Δ τ) = (g₂ τ) ^ 3 - 27 * (g₃ τ) ^ 2)) (τ : ℂ) (hτ : τ ≠ 0) :
    (Δ (-1 / τ)) = (τ / Complex.I) ^ 12 * (Δ τ) := by
      -- By the properties of Eisenstein series under modular transformations, we have:
      have h_g2 : g₂ (-1 / τ) = τ ^ 4 * g₂ τ := by
        -- Substitute $τ-1/$ into the definition of ₂$g$ and simplify the expression.
        have h_g2_sub : g₂ (-1 / τ) = (2 / 3) * ∑' (x : ℤ × ℤ), if ¬(x.1 = 0 ∧ x.2 = 0)
    then (τ ^ 4 / (x.1 * τ - x.2) ^ 4) else 0 := by
          -- Substitute $τ-1/$ into the definition of ₂$g$ and simplify the expression.
          rw [h₀]
          field_simp;
          exact tsum_congr fun x => by ring;
```

```
        rw [ h_g2_sub ];
        norm_num [ h₀, mul_assoc, mul_comm, mul_left_comm, ← tsum_mul_left ];
        rw [ ← Equiv.tsum_eq ( Equiv.ofBijective ( fun x : ℤ × ℤ => ( x.snd, -x.fst )
    ) ⟨ fun x y hxy => by aesop, fun x => by use ( -x.snd, x.fst ) ; aesop ⟩ ) ] ;
    norm_num;
        exact tsum_congr fun _ => by aesop
      have h_g3 : g₃ (-1 / τ) = τ ^ 6 * g₃ τ := by
        simp_all ( config := { decide := Bool.true } ) [ div_eq_mul_inv, mul_comm ];
        norm_num only [ mul_assoc, mul_comm, mul_left_comm, ← mul_assoc, ←
    tsum_mul_left ];
        rw [ ← Equiv.tsum_eq <| Equiv.ofBijective ( fun ( a : ℤ × ℤ ) => ( a.2, -a.1 )
    ) ⟨ fun a b hab => by aesop, fun b => ⟨ ( -b.2, b.1 ), by aesop ⟩ ⟩ ];
        refine' tsum_congr fun x => _ ; aesop;
        -- By simplifying, we can see that the two expressions are equal.
        field_simp [hτ]
        ring;
      rw [ h₂, h₂ ];
      rw [ h_g2, h_g3 ] ; ring;
      -- Since $Complex.⁻I¹ = -Complex.I$, we have $Complex.⁻I¹ ^ 12 = (-Complex.I) ^
    12 = 1$.
      have h_inv : Complex.I⁻¹ ^ 12 = 1 := by
        norm_num [ Complex.ext_iff, pow_succ' ];
      -- Substitute h_inv into the equation to simplify the right-hand side.
      rw [h_inv];
      -- Now the equation simplifies to an obvious equality, so we can conclude.
      ring
```

## D  Analysis I Examples

We include here some of the examples mentioned in Section 3 obtained testing Aristotle with [38, 39]. First, Aristotle identified four errors, now corrected [40], with the following disproofs:

```
/-- Proposition 10.1.7 (Newton's approximation) / Exercise 10.1.2 -/
theorem _root_.HasDerivWithinAt.iff_approx_linear (X: Set ℝ) (x₀ :ℝ) (f: ℝ → ℝ) (L:ℝ) :
  HasDerivWithinAt f L X x₀ ↔
  ∀ ε > 0, ∃ δ > 0, ∀ x ∈ X, |x - x₀| < δ → |f x - f x₀ - L * (x - x₀)| < ε * |x - x₀|
    := by
  sorry

--Counterexample to the above statement
example : False := by
  have := HasDerivWithinAt.iff_approx_linear
  revert this
  simp only [imp_false, not_forall]
  push_neg
  exists Set.univ, 0;
  exists 0, 0;
  norm_num [ hasDerivWithinAt_univ ]
  exact Or.inl ⟨ hasDerivAt_const _ _, 1, by norm_num, fun δ δh => ⟨ 0, by simpa using
    δh ⟩ ⟩
```

```
theorem badExample2 (x₀:ℝ) : HasDerivWithinAt (fun x ↦ (x-2)/(x-1)) (1 /(x₀-1)^2) (Set
    .univ) x₀ := by
  sorry

--Counterexample to the above statement
example : False := by
  have := badExample2
  revert this
  simp only [imp_false, not_forall]
  refine' ⟨ 1, _ ⟩ ; norm_num [hasDerivWithinAt_iff_tendsto];
  rw [ Metric.tendsto_nhds_nhds ] ; norm_num;
  use 1/2;
  field_simp;
  refine' fun x x_pos => _;
  by_cases h1 : x < 1;
  -- Case 1
  · exact ⟨ 1 - x / 2, abs_lt.mpr ⟨ by linarith, by linarith ⟩, by rw [ abs_of_neg,
    le_div_iff₀ ] <;> nlinarith ⟩;
  -- Case 2
  · exact ⟨ 1/2, abs_lt.2 ⟨ by linarith, by linarith ⟩, by norm_num [ abs_of_nonneg ] ⟩
```

```
theorem badExample3 : ¬ IsolatedPt 3 ((Set.Ioo 1 2) ∪ {3}) := by
  sorry

example : False := by
  have := badExample3
  revert this
  simp ( config := { decide := true } ) [ Chapter9.IsolatedPt ];
  exact ⟨ 1 / 2, by norm_num, fun y hy₁ hy₂ hy₃ => by cases abs_cases ( 3 - y ) <;>
    linarith ⟩
```

```
/-- Exercise 9.1.15 -/
theorem badExample4 {E:Set ℝ} (hE: Bornology.IsBounded E) : AdherentPt (sSup E) E ∧
    AdherentPt (sSup E) ᶜE := by
  sorry

example : False := by
  have := @badExample4
  revert this
  norm_num [ Chapter9.AdherentPt_def ];
  refine' ⟨ ∅, _, _ ⟩ <;> simp ( config := { decide := Bool.true } ) [ ClusterPt ]
```

For two problems, Aristotle found proofs that did not need some of the stated assumptions. Sometimes this is deliberate on the part of an author – without unnecessary hypothesis proving the statement may be too difficult for a student – but in both of these cases the assumption was expected to be necessary, and the extra assumptions were subsequently removed [40].

```
/-- Exercise 9.1.7 -/
example {n:ℕ} (hn: n ≥ 1) (X: Fin n → Set ℝ) (hX: ∀ i, IsClosed (X i)) :
    IsClosed ⋃( i, X i) := by
  --NOTE: Did not need the assumption `n ≥ 1`
  by_contra h;
```

```
    exact h ( isClosed_iUnion_of_finite fun i => hX i )
```

```
/-- Exercise 9.1.9 -/
example {X:Set ℝ} {x:ℝ} (hx: AdherentPt x X) : ¬ (LimitPt x X ∧ IsolatedPt x X) := by
  --NOTE: Did not need the assumption `AdherentPt x X`
  simp_all [ Chapter9.LimitPt, Chapter9.IsolatedPt ];
  intro h₂ h₃ ε εpos;
  obtain ⟨ y, hy₁, hy₂ ⟩ := h₂ ε εpos ; use y ; aesop
```

We also include two of the more interesting proofs that Aristotle wrote (validating the correct exercises as provable) in Chapter 4. This chapter focuses on manipulation of the Analysis I book's custom definition of the Int type as a quotient, and these solutions demonstrate Aristotle's ability to adapt to new and non-standard definitions in context.

```
/-- Lemma 4.1.11(f) (Order trichotomy) / Exercise 4.1.7 -/
theorem Int.trichotomous' (a b:Int) : a > b ∨ a < b ∨ a = b := by
  -- By definition of integers, every integer is greater than, less than, or equal to
    another integer.
  obtain ⟨a₁, a₂, ha₂⟩ := Int.eq_diff a;
  obtain ⟨b₁, b₂, hb₂⟩ := Int.eq_diff b;
  rw [ ha₂, hb₂ ];
  -- Applying the trichotomy of natural numbers to ₁$a + ₂b$ and ₂$a + ₁b$, we get three
    cases.
  have h_trichotomy : (a₁ + b₂ > a₂ + b₁) ∨ (a₁ + b₂ < a₂ + b₁) ∨ (a₁ + b₂ = a₂ + b₁)
    := by
    exact Classical.or_iff_not_imp_left.2 fun h => Classical.or_iff_not_imp_left.2 fun
    h' => by linarith;
  rcases h_trichotomy with ( h | h | h );
  · simp [Section_4_1.Int.lt_iff];
    refine Or.inl ?_;
    subst ha₂ hb₂
    simp_all only [gt_iff_lt, PreInt.instSetoid]
    apply And.intro
    · -- Since ₂$a + ₁b < ₁a + ₂b$, we can find a natural number $t$ such that ₁$a + ₂b
    = ₂a + ₁b + t$.
      obtain ⟨ t, ht ⟩ : ∃ t : ℕ, a₁ + b₂ = a₂ + b₁ + t := by
        exact Nat.exists_eq_add_of_le h.le;
      use t;
      convert Quotient.sound _;
      simp [ Setoid.r, Setoid.refl, Setoid.symm, Setoid.trans, ht ];
      ring;
    · intro a
      exact absurd ( a ) ( by simpa [ Section_4_1.PreInt.eq, eq_comm ] using by
    linarith );
  · right;
    left;
    -- By definition of formalDiff, we have ₁a + ₂b < ₂a + ₁b if and only if -(₂a - ₁a)
    < ₂b - ₁b.
    rw [Section_4_1.Int.formalDiff, Section_4_1.Int.formalDiff];
    -- Let's choose any $t$ such that ₁$a + ₂b + t = ₂a + ₁b$.
    obtain ⟨t, ht⟩ : ∃ t : ℕ, a₁ + b₂ + t = a₂ + b₁ := by
      exact ⟨ _, Nat.add_sub_of_le h.le ⟩;
```

```
    refine' ⟨ ⟨ t, _ ⟩, _ ⟩;
    · exact Quotient.sound ( by simpa using by omega );
    · by_contra h_contra;
      exact absurd h_contra ( by rw [ Quotient.eq ] ; intros H; simp_all [ Section_4_1.
  PreInt.eq, PreInt.instSetoid ] ; linarith );
  · -- If ₁$a + ₂b = ₂a + ₁b$, then $a = b$.
    right;
    exact Or.inr ( Quotient.sound <| by simpa [ add_comm ] using h )
```

```
/-- Not in textbook: equivalence preserves order and ring operations -/
abbrev Int.equivInt_ordered_ring : Int ≃+*o ℤ where  toEquiv := equivInt
  map_add' := by
    -- We'll use that the output of the function `toFun` is defined for any integer
  input.
    intro x y;
    obtain ⟨ a, b ⟩ := Int.eq_diff x;
    obtain ⟨ c, d ⟩ := Int.eq_diff y;
    obtain ⟨w, h⟩ := b
    obtain ⟨w_1, h_1⟩ := d
    simp_all only [Quotient.lift_mk]
    -- Now let's show that the equivalence relation is preserved under addition.
    have h_add : (a + c : ℤ) - (w + w_1 : ℤ) = (a - w : ℤ) + (c - w_1 : ℤ) := by
      ring;
    exact h_add
  map_mul' := by-- We'll use that the output of the function `toFun` is defined for any
   integer input.
    intro x y
    induction x using Quotient.inductionOn';
    -- By definition of multiplication in the integers, we can rewrite this equation as
  :
    -- $(a * c + b * d) — (a * d + b * c) = ((a — b) * y).1 — ((a — b) * y).2$
    have h_mul : ∀ (a b c d : ℕ), (a * c + b * d : ℤ) - (a * d + b * c) = (a - b : ℤ) *
   (c - d) := by
      intros; ring;
    induction y using Quotient.inductionOn' ;
    aesop
  map_le_map_iff' := by
    rintro ⟨ a ⟩ ⟨ b ⟩;
    apply Iff.intro;
    -- Case 1
    · rintro H;
      -- Rewrite the goal using the definition of less than or equal to in terms of the
   natural numbers.
      have h_nat : ∃ (t : ℕ), (a.minuend + b.subtrahend + t = b.minuend + a.subtrahend)
   := by
        use (b.minuend + a.subtrahend) - (a.minuend + b.subtrahend);
        rw [ Nat.add_sub_of_le ];
        contrapose! H;
        erw [ Subtype.mk_lt_mk ];
        exacts [ by simp [equivInt, PreInt.instSetoid, Quotient.lift]; omega, fun _ =>
  True, trivial, trivial ];
      -- By definition of addition in the integers, we can rewrite the goal using the
  natural numbers.
```

```
    obtain ⟨t, ht⟩ : ∃ t : ℕ, (a.minuend + b.subtrahend + t = b.minuend + a.
subtrahend) := h_nat;
    use t;
    exact Quotient.sound ( show Section_4_1.PreInt.mk b.minuend b.subtrahend ≈
Section_4_1.PreInt.mk ( a.minuend + t ) ( a.subtrahend ) from by simp [ Setoid.r, *
 ] ; linarith );
  -- Case 2
  · rintro ⟨ c, hc ⟩;
    exact hc.symm ▸ sub_le_sub_right ( Nat.cast_le.mpr <| Nat.le_add_right _ _ )
```